\documentclass{article}

\usepackage[nonatbib, final]{tackling_climate_workshop_style}

\usepackage[utf8]{inputenc} 
\usepackage[T1]{fontenc}    
\usepackage{hyperref}       
\usepackage{url}            
\usepackage{booktabs}       
\usepackage{amsfonts}       
\usepackage{nicefrac}       
\usepackage{microtype}      
\usepackage{graphicx}
\usepackage{caption}
\usepackage{amsmath}

\graphicspath{ {./images/} }

\title{Data-Driven Traffic Reconstruction and Kernel Methods for Identifying Stop-and-Go Congestion }

\author{%
 Edgar Ramirez Sanchez$^*$, Shreyaa Raghavan\thanks{These authors contributed equally to this work.} , Cathy Wu \\
  Massachusetts Institute of Technology\\
  Cambridge, MA 02139 \\
  \texttt{\{edgarrs, shreyaar, cathywu\}@mit.edu} \\
}

\begin{document}

\maketitle

\begin{abstract}
  Identifying stop-and-go events (SAGs) in traffic flow presents an important avenue for advancing data-driven research for climate change mitigation and sustainability, owing to their substantial impact on carbon emissions, travel time, fuel consumption, and roadway safety. In fact, SAGs are estimated to account for 33-50\% of highway driving externalities. However, insufficient attention has been paid to precisely quantifying where, when, and how much these SAGs take place––necessary for downstream decision making, such as intervention design and policy analysis. A key challenge is that the data available to researchers and governments are typically sparse and aggregated to a granularity that obscures SAGs. To overcome such data limitations, this study thus explores the use of traffic reconstruction techniques for SAG identification. In particular, we introduce a kernel-based method for identifying spatio-temporal features in traffic and leverage bootstrapping to quantify the uncertainty of the reconstruction process. Experimental results on California highway data demonstrate the promise of the method for capturing SAGs. This work contributes to a foundation for data-driven decision making to advance sustainability of traffic systems.
\end{abstract}

\section{Introduction}

Transportation is the most significant contributor to greenhouse gas (GHG) emissions, accounting for 28 percent of the total in the U.S.. Traffic congestion on its own is estimated to cost the average driver around 2000 USD in large cities and, to the United States alone, 305 billion USD in 2017 \cite{EPA_GHG}. “Stop-and-go”(SAG) traffic, a common phenomenon in congested traffic, happens when cars traveling on a highway periodically form waves that propagate down the highway, leading cars to constantly accelerate and brake unnecessarily.  These SAG waves or events lead to up to 67\% higher fuel consumption, longer travel times, higher GHG emissions levels, and could be a safety hazard \cite{stern2018dissipation}. Concerningly, studies have found that SAGs can be responsible for 33-50\% of highway driving externalities, which are external costs such as carbon emissions, highway accidents, etc \cite{goldmann2021quantifying,  goldmann2020economic}. To mitigate climate change and improve air quality in human settlements, making road transportation more environmentally sustainable and reducing SAG congestion should be a priority.

However, there is hope because SAGs are preventable, implying that a considerable portion of these adverse effects and externalities of highway driving are avoidable. Multiple studies have proven the potential of different technologies to alleviate SAGs using Autonomous Vehicles (AVs) \cite{balzotti2021stop, laval2010mechanism, oh2015impact, portz2011analyzing} , reinforcement learning \cite{kreidieh2018dissipating}, and variable speed limits \cite{wang2016connected}, among others. Moreover, several studies have provided insights into how these events form, propagate, and fade \cite{balzotti2021stop, laval2010mechanism, oh2015impact, portz2011analyzing}. It is clear that in theory, we can make proactive efforts to enable and support these interventions if we are able to identify when and where these SAGs occur.

In particular, machine learning could play a pivotal role, as it has the capability to identify and predict a broad range of events through techniques like image processing, kernel methods, and anomaly detection. However, a major limitation of these models is that they require rich datasets to capture the nuanced dynamics of vehicles, and there is very limited rich traffic data available. In contrast, the generally available data, of which there is a large amount, is typically too sparse or noisy to capture SAGs. In the US, for example, the highway dataset with the largest coverage is the Caltrans Performance Measurement System (PeMS), but even in this best-case scenario, sensors are spaced thousands of feet away and the data is both temporally and spatially sparse. The existing data, in its current state, is not enough to capture the minute behaviors of stop-and-go events, especially at a large scale.

In response, this paper proposes the use of traffic reconstruction as a way to fill in the gaps and generate rich trajectory information to enable the deployment of complex ML techniques. We apply this framework to a county in California and devise a kernel-based SAG identification method with bootstrapping to provide robust insights into traffic behavior. 

\section{Related Work}

The study of stop-and-go events (SAGs) in traffic has a longstanding history, dating back to the 1960s \cite{Richards_1956}. While much of the literature has traditionally focused on fundamental analyses, there has been a discernible shift toward identification methods \cite{Saxena_2016}, in part driven by challenges in obtaining sufficient data for large-scale identification. 
On one hand, theoretical studies encompass dynamic wave modeling, control mechanisms, instabilities, driving models, and other aspects that shed light on triggers, formation, propagation, and dissipation of these waves \cite{laval_characteristic, laval2010mechanism, Laval_2014, Li_2011, Wilson_2008, Chen_2014, Li_2014, Yeo_2009, Zheng_2011, Corli_2019, balzotti2021stop, oh2015impact, portz2011analyzing}. 

On the other hand, empirical studies focus on identification through various techniques, such as time series and signal processing methods, to identify SAG occurrences in space and time from the observed traffic or vehicle-specific data \cite{He_2019, Jiang_2017, Abdi_2016, Lin_2019, Saxena_2016, Li_2012, Wu_2019, Mauch_2004}. These approaches typically rely on rich trajectory data that only exists for very few stretches of highway, and only some of them have sufficient resolution to incorporate spatiotemporal features. In contrast, most studies use data from fixed and sparse sensors and identify SAGs using stationary wave processing techniques. In particular, wavelet transformation has been a popular approach for SAG identification \cite{Saxena_2016, Zheng_2011}, matching a Mexican hat profile to the signal to identify where this behavior occurs. 

Finally, while research links stop-and-go waves to negative effects such as increased fuel emissions, safety risks, and driver distress \cite{Wu_2019, Li_2014, Wismans_2015, oh2015impact}, SAG quantification has not been conducted at scale. A unique example in quantifying the number of SAGs is a study estimating that 20\% of traffic jams in the Netherlands could be stop-and-go waves \cite{noordegraaf2011filegolven}. The significance of this estimation is underscored by its inclusion as a priority in the national plan of Intelligent Transportation Technologies (ITS) \cite{Netherlands}. Yet, this is an anomaly, as most countries, including the United States, do not treat stop-and-go mitigation as a priority due to a lack of quantitative evidence.

Unlike previous literature, our work makes use of both spatiotemporal features in traffic data and broadly available data, which is sparse in nature. 

\section{Traffic Feature Identification Framework}

To leverage available yet sparse data, we propose a technique to augment raw traffic data with physics-based traffic simulations to build a richer, more continuous data representation, with the goal of enabling ML pipelines for stop-and-go identification.
\begin{figure}[h]
\centering
\includegraphics[scale=0.28]{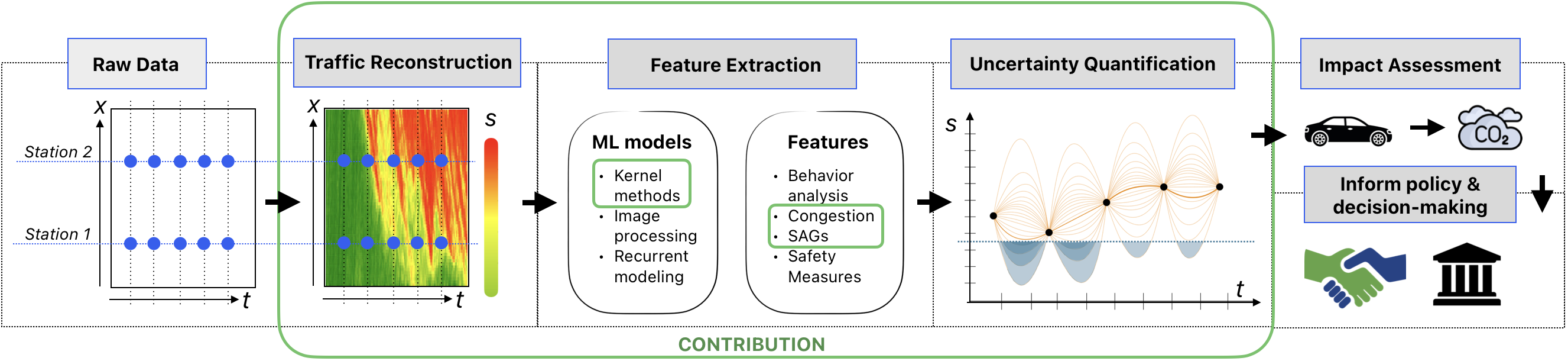}
\caption{Pipeline for traffic feature identification. Our contribution is boxed in green.}
\label{fig:pipeline}
\end{figure}
Our raw data source is the Caltrans Performance Measurement System (PeMS) dataset, which consists of roughly 39,000 induction loop detectors (electromagnetic highway sensors) that capture traffic flow, occupancy, and speed aggregated every 30 seconds on most freeways in major metropolitan areas of California. The key concern with PeMS is that the dataset is too sparse to draw conclusions about traffic events or have a complete understanding of vehicle trajectories. As a result, we seek to fill in these data gaps via traffic reconstruction while using PeMS as our ground truth data.

\subsection{Traffic Reconstruction}
We leverage a traffic reconstruction model built as part of a safety analysis framework, which simulates traffic scenarios at a fine-grain level to build individual vehicle trajectories. It does so by combining micro-level dynamics of driving behavior on California highways, captured by highway video data, with macro-level data obtained by induction loop detectors.  

At the large scale, we represent the entire road network as a graph $G = (E, V)$. The goal is to recover the optimal flow through every edge $e$ at time $t$ such that the routes of the vehicles align closely to the ground-truth traffic data while still being realistic. We model this flow estimation problem as a maximum flow problem. At the smaller scale, we create a driver model that relies on the widely used “Intelligent Driver Model” (IDM) \cite{treiber2000congested}. In our traffic reconstruction framework, the driver model is calibrated to digital video footage of California highways to better capture behaviors commonly seen in that region \cite{zhang2022bayesian}. Both the driver model and the estimated vehicle routes are then combined in Simulation of Urban MObility (SUMO), a traffic simulator, to extrapolate detailed trajectories of traffic. The outcome is a rich representation of an entire highway network, that has information about every vehicle at every point in space and time. A more comprehensive overview of this traffic reconstruction method is currently in preparation.

\subsection{Uncertainty Quantification}
With a continuous, time-space representation of traffic, we can now extract salient features. However, each reconstruction of traffic may have deviations, so relying on a single reconstruction for predictions could inflate or erase certain behaviors. This prompts us to question: given a single simulation, how certain are we in an observed event or behavior? Instead, we want a robust measure of confidence in our identification of features. 

To accomplish this, we use a technique similar to bootstrapping and leverage our traffic reconstruction simulator as a distribution sampler, which allows us to estimate the distribution of speeds for a point in time and space \cite{efron1994introduction}. We replicate the traffic reconstruction $k$ times and build a distribution of average vehicle speed at position $x$ and time $t$: $p_s(s | t, x)$. In each replication, we randomly sample (with replacement) stochastic elements of the reconstruction, such as speed deviations, offset in departure times, and vehicle driving dynamics.

Then, we define an event or feature of interest, $A$, such that $A(t, x) = 1$ if $A$ occurs at $(t, x)$, and $A(t, x) = 0$ otherwise. $A$ is a function of speeds, $S_{t,x}$, such that $A(t, x) = f(S_{t, x})$.  $S_{t,x} = \{s_{t+i, x+j}$  :  $ \forall i \in [-m, m], j \in [-n, n]\}$ is a neighborhood of speeds surrounding $s_{t, x}$, where $m$ and $n$ are parameters that control how wide this neighborhood is. To provide a measure of certainty, we want to know the probability of $A$ occurring at $(t, x)$, which is $p_A(A | t, x)$. Since $A_{t, x} = f(S_{t,x})$, $ p_A(A | t, x) = p_s(f(S_{t, x}, ..., \nu) | t, x) $. Hyperparameters for $f$ are denoted as $\nu$.  This function $f$ and collection $S_{t,x}$ differ based on the setting and behaviors of interest. 

\section{Application to Stop-and-Go Event Identification}

Using the individual trajectories from the traffic reconstruction, we build a rich time-space representation of traffic that encodes meaningful features. This representation, known as a time-space diagram, is commonly used in the transportation community to visualize vehicle behaviors as they move through time (x-axis) and space (y-axis). The color of the diagram encodes the aggregated speed of the vehicles across all lanes. In our study, we reconstruct 100 variations of traffic for the entire Los Angeles County from 11:15 to 11:30 a.m.. We focus our analysis below on a 0.5 km segment of the 110-N freeway. 
\begin{figure}[h]
\centering
\includegraphics[scale=0.29]{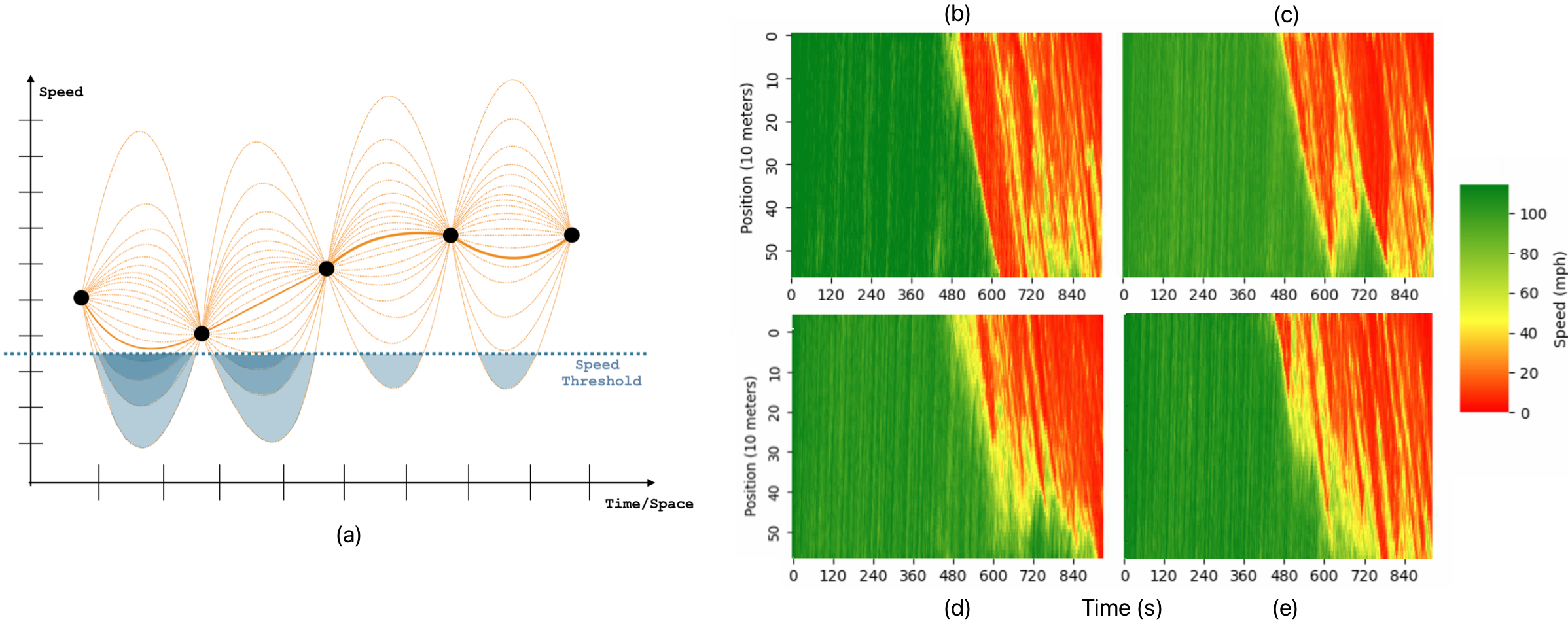}
\caption{(a) Schematic of possible reconstructed trajectories that fill the gaps between ground-truth readings. These trajectories vary in each replication. (b-e) 4 varying time-space diagrams out of 100.}
\label{fig:sims}
\end{figure}

A stop-and-go event is defined by an alternating wave of deceleration and acceleration, or high speeds and low speeds. Despite the minimal work done on identification specifically, extensive research has focused on explaining and characterizing this oscillating pattern using fundamental traffic models, such as the Lighthill–Whitham–Richards (LWR) model \cite{hystersis, oscillation}. From the literature, we characterize a stop-and-go event as a high $\rightarrow$ low $\rightarrow$ high speeds profile that may or may not have subsequent SAGs. In a time-space representation, they appear as waves with a negative slope, traveling backwards in space because of vehicles slowing down behind the current SAG and progressively moving its center backwards \cite{laval_characteristic} . This behaviors can be seen in Figure 2(b-e). Additionally, empirical studies have recorded concrete properties of SAGs: for example, during rush hour on US-101, the average width of a SAG was recorded to be 7 min with an average amplitude of 15 km/hr \cite{laval_characteristic}. We capture these known behaviors with a parameterized spatiotemporal kernel.

We take into account both the dip in speed over time and the backward movement over space by designing a 2-D kernel based on the Sobel operator for diagonal edge detection \cite{kanopoulos1988design}. However, the behavior of a SAG depends on many factors such as the time of day or road conditions, so not all SAGs appear the same on a time-space diagram. To capture SAGs with different oscillation periods, we parameterize the width of the kernel. It is important to note that other properties of the kernel, such as the slope of propagation or the magnitude of the dip (which corresponds to the amplitude of the wave) can also vary depending on the setting. Here, we only specify the width, i.e. the wave period, as a parameter.

An example of the kernel $K$ with width $=4$ seconds is as follows:
\[
K(4)=
  \begin{bmatrix}
    0 & -1 & -1 & -1 & -1 & 0 & 2 & 2 \\
    2 & 0 & -1 & -1 & -1 & -1 & 0 & 2  \\
    2 & 2 & 0 & -1 & -1 & -1 & -1 & 0\\
  \end{bmatrix}
\]

We convolve this kernel $K(w)$ across neighborhoods of speeds, $S_{t, x}$, in the time-space diagram and normalize these values from $-1$ to $1$. $S_{t,x}(m,n) = \{s_{t+i, x+j}$  :  $ \forall i \in [-m, m], j \in [-n, n]\}$ is a neighborhood of speeds surrounding $s_{t, x}$. We define this convolution as the function $f_{SAG}(S_{t, x}, w |t, x) = norm_{(-1,1)}(K(w) \ast S_{t, x}(w, 1))$, which we also refer to as the "Kernel Activation Value". The Kernel Activation Value provides a measure of how strongly the behavior at $(t, x)$ matches the kernel. 

Then, we define $C_{t,x}$ as the event that a SAG occurs at $(t,x)$, which is 1 when  $f_{SAG}(S_{t, x}, w |t, x) \geq \epsilon$ and $0$ otherwise. This $\epsilon$ controls how steep the stop-and-go behavior should be in order to be considered a SAG event. For our study, we use a threshold of $\epsilon = 0.30$. We build an indicator variable $C_{t, x}^{j}$ defined by the occurrence of a SAG in trial $i$ at $(t,x)$ 

Then, we estimate the probability of a SAG occurring at $(t, x)$ by using the $C_{t, x}^{j}$ as samples of the probability distribution:$$p( C_{t,x} = 1|t, x) = p_s(f_{SAG}(S_{t, x},w |t, x) \ge \epsilon | t, x) = \frac{1}{100} \sum_{i=1}^{100} C_{t, x}^{i}(w, \epsilon) $$
Using the continuous time-space representation and our kernel-based method, we are able to detect and visualize stop-and-go traffic in time and space, as shown by the blue diagonal lines in the first row of Figure 3. The probabilities in the second row of Figure 3 create hotspots that we know are more prone to SAGs. As a result, we can quantify how certain we are that a SAG detected in a single trial is not simply variance or noise. This probabilistic analysis allows us to extrapolate the observed traffic patterns in data to the real world and provides a measure of uncertainty, which is crucial for real-world decision-making and intervention.  
\begin{figure}[h]
\centering
\includegraphics[scale=0.40]{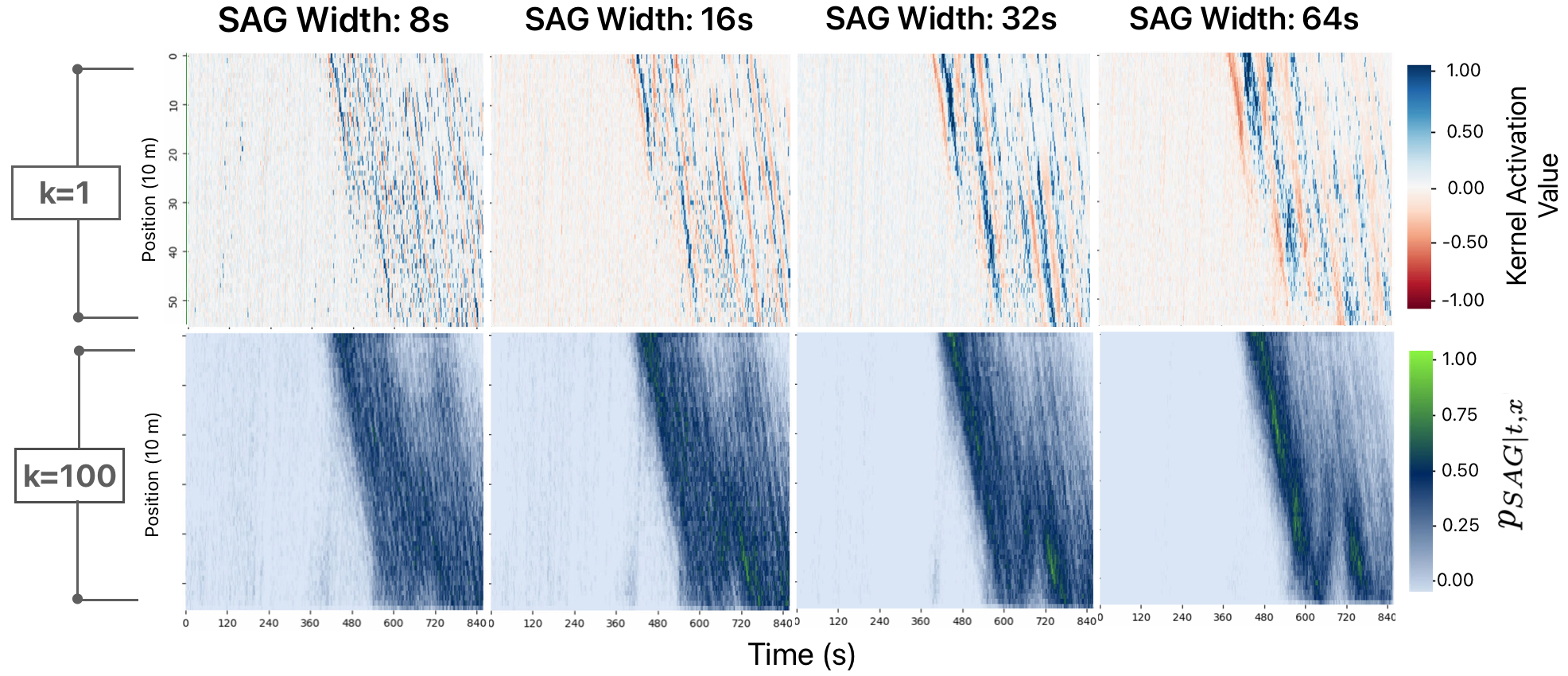}
\caption{$f_{SAG}$ for kernels of different widths applied to a single trial (same trial as Figure 2(b)) are shown in the first row. Values classified as SAG ( $ > \epsilon$) are boosted for visual reference. The second row shows the probability (combined across all 100 simulations) of a SAG occurring. $U$ represents the percentage of the heatmap for which the detector is uncertain, i.e. $0.25 < p_{SAG} < 0.75$.}
\label{fig:sims}
\end{figure}

\section{Conclusion}

Stop-and-go wave mitigation is a significant opportunity for making highway driving more sustainable. This paper presents a novel traffic reconstruction framework, offering a rich spatio-temporal representation that effectively and robustly detects traffic behaviors, particularly stop-and-go events. Our visual representation enables the creation of comprehensive traffic datasets, while the success of our kernel method in capturing SAG patterns demonstrates the potential for more complex ML methods to leverage this data. With local SAG identification, policy-makers can pinpoint when and where SAGs are likely to occur in their communities, empowering the design and deployment of targeted interventions. Through SAG identification at scale, we can quantify how much of the total highway congestion is avoidable. This would concretely motivate the need for smart infrastructure and other technology-based interventions and is a crucial step toward convincing decision-makers to take action at a national level.

Future research directions include the development of sequential ML tools for time-series analysis, extending to other features such as safety measures, causal analysis via individual trajectories, behavior analysis, and prediction through supervised learning. A key goal in the future will be to validate the traffic reconstruction more quantitatively. We plan to compare reconstructed trajectories to the available ground truth by testing the pipeline on a richer dataset, such as a vehicle trajectory dataset or highway camera footage. Finally, scaling the SAG identification process for the full California highway system and for longer periods of time is a promising step towards at-scale SAG intervention and more sustainable roadways.

\bibliographystyle{ieeetr}
\bibliography{tackling_climate_workshop_style}

\end{document}